\documentclass[conference]{IEEEtran}
\IEEEoverridecommandlockouts
\usepackage{cite}
\usepackage{amsmath,amssymb,amsfonts}
\usepackage{algorithmic}
\usepackage{graphicx}
\usepackage{textcomp}
\usepackage{xcolor}
\usepackage{balance}  
\usepackage{tabularray}
\usepackage{adjustbox}
\usepackage{multirow}

\def\BibTeX{{\rm B\kern-.05em{\sc i\kern-.025em b}\kern-.08em
    T\kern-.1667em\lower.7ex\hbox{E}\kern-.125emX}}
\begin{document}

\title{Autonomous Subsea Cable Search and Tracking with Graph-Optimised Priors and Visual Tracking\\
\thanks{This project is funded by the Defence and Security Accelerator (DASA) Project ID ACC2037839. The Smarty200 AUV was developed under the EPSRC equipment grant EP/V035975/1 with support from SMMI, IROE and IRIS.
The subsea communication cables used in our field trials were provided by the British Telecom Group. Susan Gourvenec is supported by the Royal Academy of Engineering through the Chairs in Emerging Technologies Scheme}
}

\author{
    \IEEEauthorblockN{
        Ibrahim Fadhil Djauhari\textsuperscript{1},
        Adrian Bodenmann\textsuperscript{1},
        Samuel Simmons\textsuperscript{1},
        Cailei Liang\textsuperscript{1},\\
        David White\textsuperscript{1},
        Susan Gourvenec\textsuperscript{1},
        Tom Bennetts\textsuperscript{2},
        Darryl Newborough\textsuperscript{2},
        Blair Thornton\textsuperscript{1,3}
    }
    \vspace{3pt}
    \IEEEauthorblockA{
        \textsuperscript{1}Southampton Marine \& Maritime Institute, University of Southampton, UK\\
        \textsuperscript{2}Sonardyne International Ltd., Yately, UK\\
        \textsuperscript{3}Institute of Industrial Science, The University of Tokyo, Japan
    }
}

\maketitle
\begin{abstract} 
Global communications rely on subsea cable infrastructure that remains vulnerable to damage from natural hazards and human activity. Autonomous underwater vehicles (AUVs) offer an efficient means to inspect long sections of exposed cable, but uncertainty in cable route maps, small cable diameters and partial burial makes continuous tracking a challenge. This paper presents a novel cable search and tracking method that leverages uncertain prior cable route maps. Graph-based optimisation continuously update the cable route to remain consistent with visual observations. Route uncertainty is constrained as a function of distance from observations using physics-based catenary models that account for cable parameters (i.e., lay depth, diameter, and density), bounding the search space to physically feasible regions  and improving search efficiency. Cable detection is performed using a semi-supervised classifier running in real-time on-board a camera-equipped AUV. These detections both update the graph-based optimisation and enable visual cable tracking. When tracking is lost due to misclassification, burial or imperfect control, the bounded search space enables efficient recovery. The approach was demonstrated in field trials using the University of Southampton's Smarty200 AUV. The system successfully located the cable despite deliberate errors in it initial cable route map, updating this to be consistent with observations and using visual tracking to inspect up to 59\% of a 120\,m test cable, with successful recovered after tracking loss. 
\end{abstract}


\section{Introduction}

\noindent Subsea cables are a vital for global communications. Over 95\% of international data passes through subsea cables~\cite{Clare2023} that span a total of over 1.4 million kilometers~\cite{TeleGeography2024}. However, subsea cables are vulnerable to damage, with over 100 subsea cable failures each year caused by both natural hazards such as earthquakes and landslides, wear, and human activities such as fishing and anchoring~\cite{SubmarineTelecoms2024, cable_breakage}. Although cables are typically buried in shallow water up to 1.5\,km depth~\cite{Douglas2017}, cables beyond this depth and in areas of hard substrates typically lay exposed on the seafloor. Furthermore, initially buried cables can become exposed over time due to seabed processes such as sediment transport and scour~\cite{Kogan:2006:templ}. 

Autonomous Underwater Vehicles (AUVs) have the potential for efficient subsea cable inspection due to their long range and   hovering capabilities. However, reliable cable following remains challenging due to small cable diameters (typically less than 3\,cm) and uncertainty in prior route knowledge, which is often on the order of 5–10\% of the lay depth due to currents during installation~\cite{MakaiLay:2022:templ}. Furthermore, cables can move over time due to natural phenomena or human activities, in some cases displacing $15-975$\,m from their initial position. Combined with AUV localisation uncertainty and the limited swath of the high-resolution sensors required for cable detection, these factors render traditional waypoint following ineffective, necessitating real-time detection to guide navigation for reliable inspection.

Optical methods can achieve sufficient realtime resolution to resolve cables and determine their relative position and orientation, where this information can be used for AUV navigation. While alternatives such as active electro-magnetic sensing can detect cables (includes buried sections) from up to 5\,m range with $\sim$10\,m measurement footprint~\cite{Szyrowski2013-2}, they indicate bulk presence of ferrous material, which is not ideal for guiding navigation. High-resolution acoustics such as Synthetic Aperture Sonar (SAS) can achieve large swaths ($\sim$100\,m) at sub-cm-resolution, but require significant onboard processing to achieve this for real-time AUV navigation, and has limited capacity to detect abrasion and other forms of damage.

Visual inspection of subsea cables and pipelines has traditionally been performed by remotely operated vehicles (ROVs). ROVs provide precise tracking through direct operator control but are tethered and require a dedicated support vessel, limiting their scalability for routine deep-water surveys. More recently, AUV-based cable and pipeline tracking methods have been developed using electromagnetic guidance \cite{Inzartsev2008, Xiang2016}, vision-based undersea telecommunication cable tracking using particle filters \cite{Ortiz2011}, and multi-sensor architectures integrating sonar and vision \cite{Evans2003, Zhang2017}. These methods perform reactive cable tracking, where the AUV follows the cable while it remains detectable but do not maintain or update a global estimate of the cable route. 

This research develops a method that, given an approximate cable route map (or cable prior) enables a camera equipped AUV to search for and track exposed subsea communication cables while maintaining a global estimate of the full cable route, including unobserved sections. The approach employs graph-based optimisation, commonly used in Simultaneous Localisation and Mapping (SLAM)~\cite{Thrun2006} to refine estimates of a robot's path to be consistent with environmental observations. In contrast, our method inverts this concept: graph-based optimisation refines estimates of the full cable route to so that they remain consistent with observations made from robotic poses, enabling updated predictions of unobserved cable sections to guide search-based path planning. The uncertainty of these predictions is bounded using physics-based catenary models to constrain the search-space around predictions to physically feasible regions. Camera-based cable detection is achieved using real-time classifiers~\cite{geoclr_2022}, which generate tracking vectors for the AUV when the cable is visible. The approach is robust compared to alternatives, as observation-updated global cable route estimates continue to guide AUV search even when the cable is not detected. This paper presents the method and demonstrates its performance during field trials conducted on subsea communication cables laid on the seafloor. 


\section{Method}

\begin{figure}[!htb]
    \centering
     \includegraphics[width=0.9\linewidth]{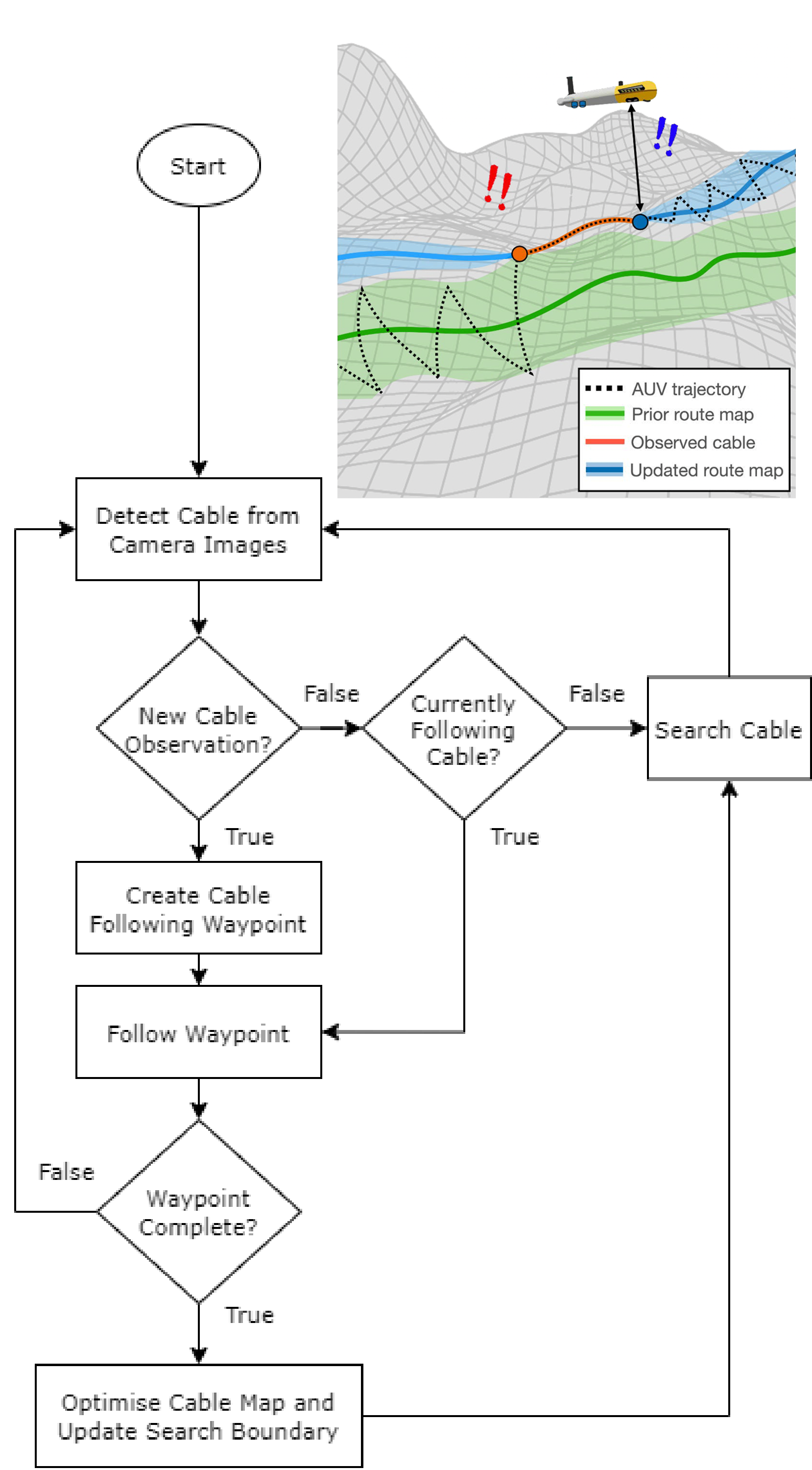}
    \caption{Flowchart illustrating the components of the cable tracking system. When the cable is detected in an image, the system generates an observation and updates waypoints to follow the cable direction. If the robot loses track of the cable, it optimises the global cable route map and updates the search area using catenary models that expand with distance from observations, accounting for cable physics. The robot continues searching until the cable is detected again, and the process repeats.}
    \label{fig:cable_tracking_flowchart}
\end{figure}

\noindent The subsea cable tracking system consists of the following parts: cable search, cable following, and global cable route estimation, as illustrated in Fig. \ref{fig:cable_tracking_flowchart}. 

\subsection{Cable-relative Coordinate Frame}

\begin{figure}[!t]
    \centering    
    \includegraphics[width=1\linewidth]{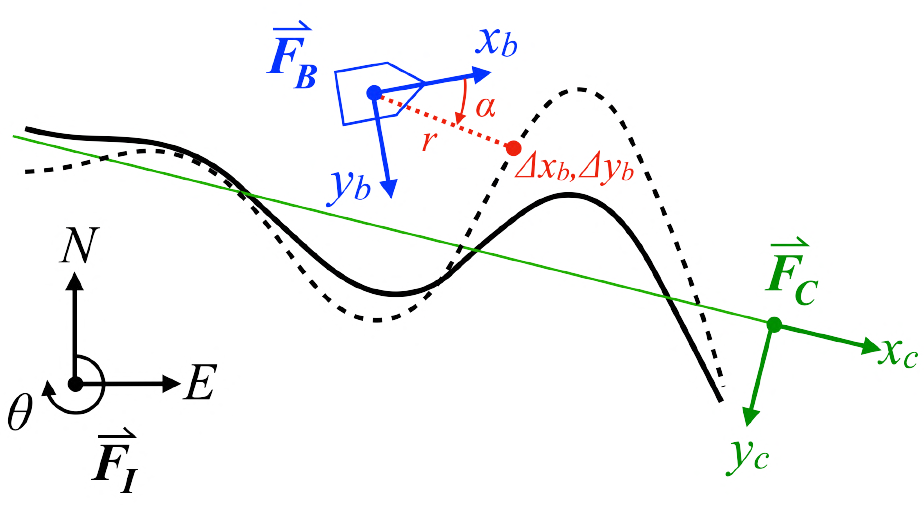}
    \caption{Illustration of the coordinate frames used in the subsea cable tracking system: the environment frame $\vec{F}_I$, AUV body frame $\vec{F}_B$, and cable frame $\vec{F}_C$, all following the NED convention. The dotted curve represents the true cable, which may deviate from the prior map (solid curve) while maintaining a similar principal axis. A cable segment is observed at range $r$ and bearing $\alpha$, located at $(\Delta x_b,\Delta y_b)$ relative to the AUV position $(x_b,y_b)$.}
    \label{fig:cable_coordinate}
\end{figure}

Subsea cables look identical along their length. In order to match featureless cable observations along the cable route map, we introduce a novel cable-relative coordinate system. This is illustrated in Fig. \ref{fig:cable_coordinate}, which shows the cable-relative coordinate frame consisting of the $along$ ($x_c$) and $normal$ ($y_c$) axes, representing the distance along the cable length and the lateral displacement of the cable's position, respectively. This coordinate frame is defined by fitting a linear regression through nodes at a regular interval that are defined along the cable prior over the planned survey distance, with the $x_c$ along the regression in the direction of increasing cable node index; the normal axis $y_c$ defined perpendicular to it, with positive values following the North-East-Down (NED) convention. This cable-relative coordinate frame solves the correspondence problem, allowing featureless cable observations to be correlated to specific points along the cable length. 

This linear frame assumes that the cable normal offset is unique with distance along $x_c$, which is reasonable considering that cable routes are typically laid to minimise the cable length required to connect distance shore landing stations. In regions where there is significant curvature of the prior (e.g., to avoid known hazards or protected areas, or near-shore landing zones), the cable prior can be split into shorter approximately linear segments that satisfy this assumption. For cables with more complex geometry, arc-length parameterisations could be introduced along the cable's centre of curvature, though this has not been implemented yet in our work.

Coordinate transformations between the environment frame $\vec{F}_I$, AUV body frame $\vec{F}_B$, and cable frame $\vec{F}_C$ are handled using $SE(2)$ homogeneous transformation matrices. The cable coordinate in $\vec{F}_C$, seen from the environment frame $\vec{F}_I$, is expressed as $H_{IC}$, and its inverse gives the transformation from the environment to cable coordinates:
\begin{equation}
    H_{CI} = H_{IC}^{-1} = \begin{bmatrix} R_{IC}^T & -R_{IC}^T t_{IC} \\ 0 & 1 \end{bmatrix},
    \label{eq:hci}
\end{equation}

\noindent where $R$ and $t$ represent the corresponding rotation matrix and translation vector. Cable observetions are made relative to the AUV body frame $\vec{F}_B$ with a translational offset $t_{BB'}$, and the AUV's pose relative to frame $\vec{F}_I$ is expressed as $H_{IB}$. The observed cable can be located in the cable coordinate frame as:
\begin{equation}
    \begin{bmatrix} t_{CB'} \\ 1 \end{bmatrix} = H_{CI} H_{IB} \begin{bmatrix} t_{BB'} \\ 1 \end{bmatrix},
\end{equation}
 
\noindent where:
 
\begin{equation}
    t_{BB'}=\begin{bmatrix} \Delta x \\ \Delta y \end{bmatrix}=\begin{bmatrix} r\cdot cos(\alpha) \\ r\cdot sin(\alpha) \end{bmatrix}.
\end{equation}

\noindent where $r$ and $\alpha$ are the range and bearing of the cable observation from the AUV. 

\subsection{Cable Detection and Cable Following}

Real-time visual tracking is achieved using an onboard model to detect sections of cables in AUV camera imagery and use these to generate a waypoint vector. The model is pre-trained using semi-supervised learning method described in our previous work~\cite{geoclr_2022}. Each image is colour corrected and split into $N\times N$ tiles, and the model detectstiles that contain cable~\cite{auv2024}. 

A cable observation is defined as a pose $(x_{obs}, y_{obs})$, bearing $\beta_{obs}$, and the index of the cable pose corresponding to the observation $i_{obs}$. Fig. \ref{fig:cable_detection} illustrates the approach for $N=3$, where an observation requires at least two tiles with positive detections for the bearing to be calculated. After an observation is made, a cable following waypoint is generated for the AUV to follow. The position of the waypoint is formulated as:

\begin{figure}[!b]
	\centering
	\includegraphics[width=0.85\linewidth]{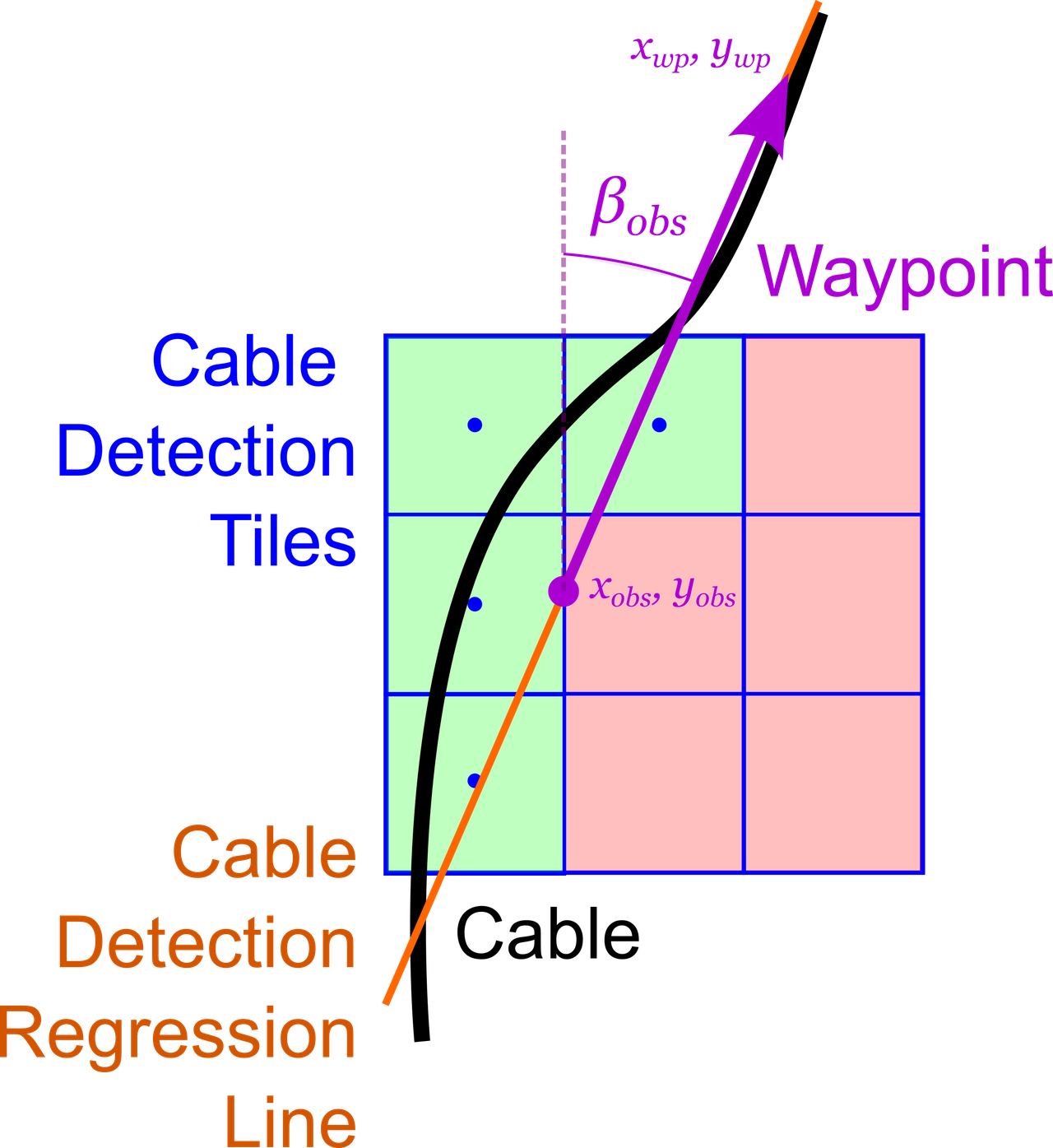}
	\caption{Illustration of $3\times3$ cable detection tiles. Green tiles indicate cable detections, while red tiles indicate no detection. The observation position $(x_{obs},y_{obs})$ is the mean of the detected tile positions. The bearing $\beta_{obs}$ is obtained by fitting a line to the detected tiles using linear regression, measured relative to the along-cable axis. A cable-following waypoint is generated as a vector of length $L_{wp}$ from $(x_{obs},y_{obs})$ to $(x_{wp},y_{wp})$. All quantities are use the cable coordinate frame, with the $x_c$ and $y_c$ directions shown in the top-left corner.}
	\label{fig:cable_detection}
\end{figure}

\begin{equation}
    x_{wp}=x_{obs}+L_{wp} \cdot cos(\beta_{obs})
\end{equation}
\begin{equation}
    y_{wp}=y_{obs}+L_{wp} \cdot sin(\beta_{obs}),
\end{equation}

\noindent where $(x_{wp}, y_{wp})$ is the waypoint coordinate in the cable coordinate frame and $L_{wp}$ is the cable-tracking look-ahead distance. Fig. \ref{fig:cable_detection} illustrates this, where we assume that the next cable section will be somewhere within a vector between the observation $(x_{obs},y_{obs})$ and the waypoint $(x_{wp}, y_{wp})$. Path following is implemented so that the AUV follow the path between the observation and the waypoint to minimise the impact of currents, where a Line-of-sight (LOS) method~\cite{fossen_2011} is used to minimise cross-track error.

\subsection{Graph-based Cable Map Optimisation}




The method uses graph-optimisation to update an estimated cable route map based on cable detections. The cable route map is initialised using subsea cable map coordinates that are logged during cable lay operations. This forms a prior that we expect to be accurate within 5 to 10\% of the cable lay depth~\cite{MakaiLay:2022:templ}. Cable coordinates in $\vec{F}_I$ that fall within the planned survey area are used to establish $\vec{F}_C$ based on linear fitting, and the mapped points are sampled to $n$ evenly spaced points along the $x_c$ axis to define the prior nodes of the graph $\mathbf{x}_i=[x_{c,i},y_{c,i}]^T$ for $i=0,...,n$. The graph's state vector $\mathbf{X}$ is initialised as the set of $2 n \times 1$ cable prior coordinates.

Every time a cable is observed by the AUV, it's location in the image $\mathbf{z}_{jt}$ is projected to the cable coordinate frame based on the AUV's pose $\mathbf{b}_{t}=\{x_{b,t},y_{b,t}, \theta_{b,t}\}$ to create the observation node $\mathbf{m}_j$, which is added to the graph as $\mathbf{m}_j =[x_{obs,j},y_{obs,j}]^T$, for the cable observation index $j=0,...,l$, where $l$ is the total number of cable observations. Correspondences are established as edges between cable nodes $\mathbf{x}_k$ and neighbouring observations that are interpolated to determine $y'_{\mathrm{obs},j'}$ that aligns with the along-track cable node at $x_{c,k}$. The graph's state vector is defined as follows:
\begin{equation}
    \mathbf{x}_i = \begin{bmatrix} x_{c,i} \\ y_{c,i} \end{bmatrix}, \quad \mathbf{m}_j = \begin{bmatrix} x_{obs,j} \\ y_{obs,j} \end{bmatrix}, \quad \mathbf{X} = \begin{bmatrix} \mathbf{x}_0 \\ \vdots \\ \mathbf{x}_n \\ \mathbf{m}_{0} \\ \vdots \\ \mathbf{m}_{l} \end{bmatrix}.
\end{equation}


\begin{figure}[!b]
    \centering
    \includegraphics[width=1\linewidth]{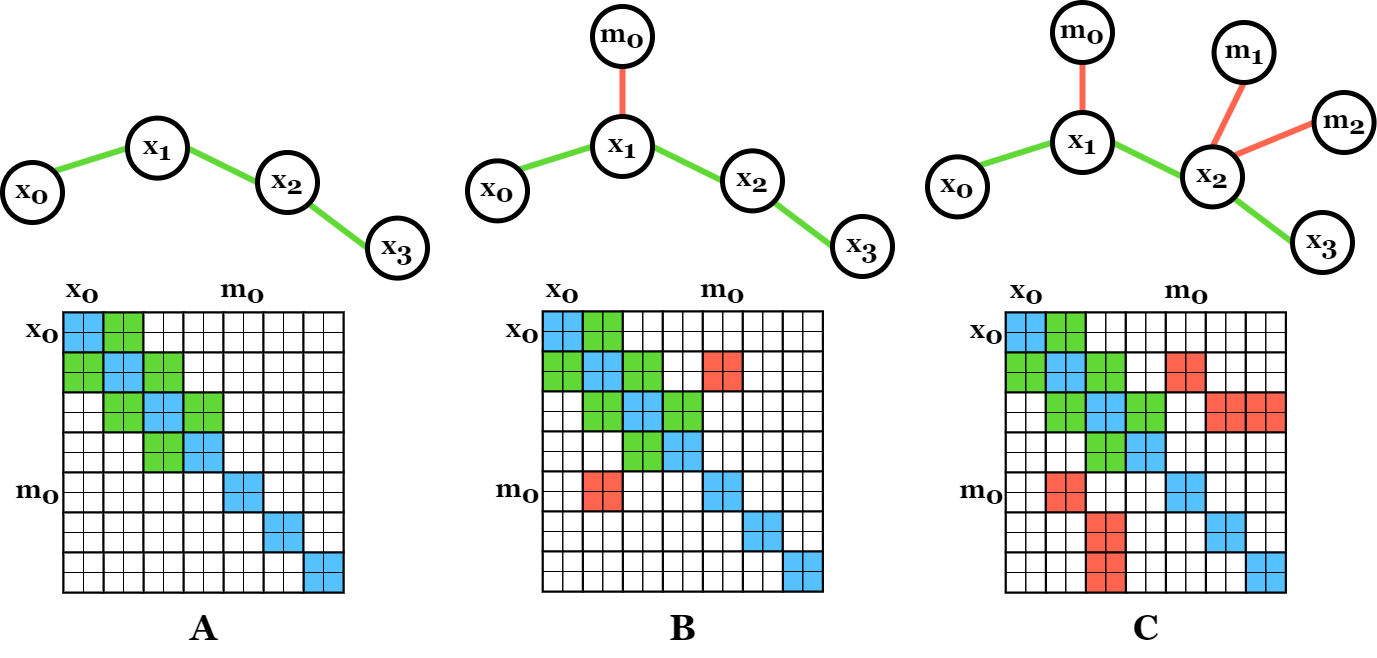}
    \caption{Illustration of graph construction with its corresponding information matrix $\Omega_{graph}$. Each cable position (blue) is connected by a motion constraint (green) to its subsequent position and by a measurement constraint (red) to the nearest observation. Graph A shows the full set of initialised cable positions. In Graph B, a new observation $\mathbf{m}_0$ is added and linked to the closest cable position $\mathbf{x}_1$. In Graph C, two new observations $\mathbf{m}_1$ and $\mathbf{m}_2$ are introduced, each connected to their nearest cable position $\mathbf{x}_2$.}
    \label{fig:graph_slam}
\end{figure}

\noindent Fig.~\ref{fig:graph_slam} illustrates the front end graph construction of the $[2(n+l)\times 2(n+l)]$ information matrix. Each cable coordinate $\mathbf{x}_i$ is connected to the next position $\mathbf{x}_{i+1}$ by an edge that constrains the relationship between the two nodes. The error between $\mathbf{x}_i$ and $\mathbf{x}_{i+1}$ is formulated as:

\begin{equation}
    e_{\mathbf{x}_i} = (\mathbf{x}_{i+1} -\mathbf{x}_i)-(\mathbf{x}_{0,i+1} -\mathbf{x}_{0,i}), \label{eq:error}
\end{equation}

\noindent where $\mathbf{x}_{0,i+1}$ and $\mathbf{x}_{0,i}$ denote the initial positions of the nodes at indices $i+1$ and $i$. Initially, these relative displacements are equal, resulting in an error of $e_{\mathbf{x},i} = 0$. During graph optimisation, the cable position in the left-hand term of \eqref{eq:error} changes, while preserving the prior cable shape and ensuring smooth variation between observations. AUV poses are not optimized within the graph, instead the measurement error minimized by the solver is the translational offset between the observed cable node $\mathbf{m}_j$ and its corresponding cable position $\mathbf{x}_k$:

\begin{equation}
    e_{\mathbf{m}_{jk}} = \mathbf{m}_j - \mathbf{x}_k.
\end{equation}

\noindent The total error of the graph can be formulated in terms of the error $e_{\mathbf{x}_i}$ and measurement error $e_{\mathbf{m}_{jk}}$:

\begin{equation}
    F(\mathbf{X}) = \sum_{i=0}^{n-1} e_{\mathbf{x}_i}^T \ \mathbf{\Omega}_{Q_i} e_{\mathbf{x}_i} + \sum_{j=0}^{l}  e_{\mathbf{m}_{jk}}^T \mathbf{\Omega}_{R_{jk}} e_{\mathbf{m}_{jk}},
\end{equation}

\noindent where $\mathbf{\Omega}_{Q_i} = Q_i^{-1}$ and $\mathbf{\Omega}_{R_{jk}} = R_{jk}^{-1}$ are the information matrices for the process and measurement, respectively. The process covariance matrix $Q_i$ represents the spatial prior uncertainty of the cable structure itself, modeling the maximum expected lateral shift between node segments derived dynamically from the catenary calculation constraints after initialisation. The measurement covariance matrix $R_{jk}$ represents the uncertainty in the observation. Assuming camera calibration errors are negligible in comparison, $R_{jk}$ is dominated by the AUV's absolute positional uncertainty. Each edge constraints acts like a spring, pulling on the nodes based on how certain each edge.

The objective of the optimisation is to estimate the arrangement of cable route and observation nodes
$\mathbf{X^*}$ that minimises the total error $F(\mathbf{X})$:

\begin{equation}
    \mathbf{X}^* = \arg \min_{\mathbf{X}} F(\mathbf{X}).
\end{equation} 

\begin{figure}[!b]
    \centering
    \begin{minipage}[b]{\linewidth}
		\centering
        \includegraphics[width=1\linewidth]{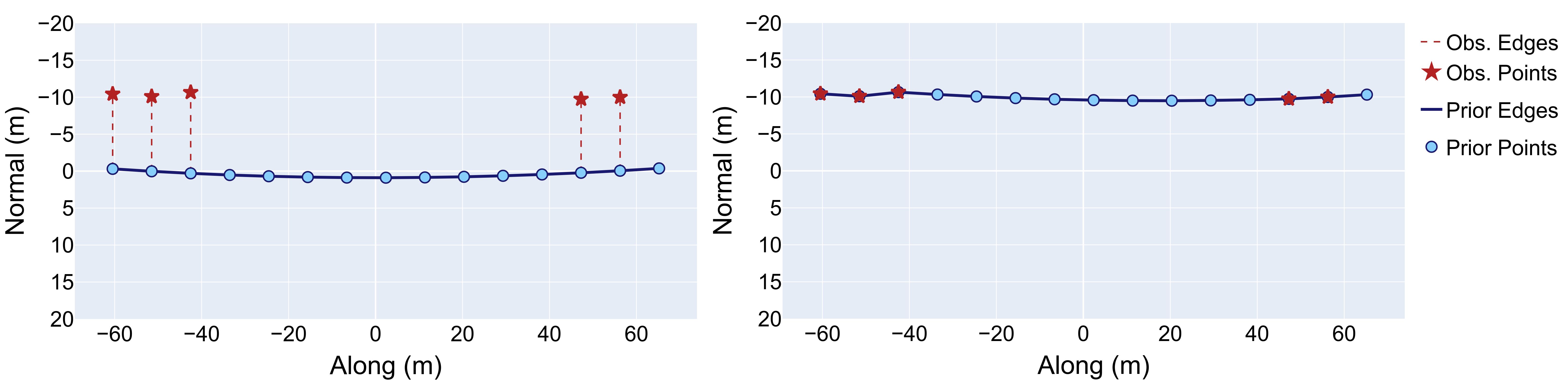}
	\end{minipage}
    \caption{A cable route map shown before (left) and after (right) optimisation based on five observations. Since observational uncertainties are small compared prior cable route map information, the cable nodes move near their corresponding observations after map optimisation. Even though only five cable positions are linked to the observations, the entire cable map is updated to be consistent with the observations and the original shape of the cable is largely retained, varying only smoothly between the observations.}
    \label{fig:optimisation_update}
\end{figure}

The optimisation is solved using the Levenberg-Marquardt algorithm implemented via the g2o library~\cite{g2o}. It is assumed that the uncertainty of the cable observations $R_{jk}$ is small relative to the prior uncertainty of the cable map positions $Q_{i}$, which is not unreasonable using established AUV navigational suites consisting of a Doppler velocity log, true-north seeking gyros and ultra-short baseline acoustic positioning~\cite{Paull2014}. 
Therefore, the cable route map at $\mathbf{x}_k$ will moved to near its corresponding observation $\mathbf{m}_j$ while still retaining the initial overall shape of the cable map due to its spatial constraint with the next position $\mathbf{x}_{k+1}$. An example showing the cable map positions before and after a map optimisation is shown in Fig. \ref{fig:optimisation_update}.

Since our method relies on featureless matching of observed cables to their location in a map, false positive detections can cause irreparable damage to the cable route estimate. 
To address issues with imperfect cable detection, we introduce a consecutive observation filter before data association. This filter takes into account the per-frame false positive rate of the cable detection method, which has been characterised in our previous work~\cite{auv2024}. An edge is only created if a set of consecutive cable observations satisfies the following spatial and a temporal criteria:
 
\begin{itemize}
    \item \textbf{Spatial:} The distance between consecutive cable detections $j$ and $j+1$ is less than the distance threshold $d_{obs}$.
    \item \textbf{Temporal:} The set of consecutive detections exceed a minimum of $N_{obs}$ consecutive observations.
\end{itemize}
 
The values of $d_{obs}$ and $N_{obs}$ are selected based on the AUV's forward speed and cable detection rate, where we assume the AUV is in continuous forward motion. Once a set of observations satisfies these criteria, the observations are associated with the corresponding cable route node based on the distance along the cable's major axis $x_c$, using linear interpolation using neighbouring observation as previously described. This filter does not account for spatially correlated false positives from continuous linear seafloor features such as anchor drag marks, ropes, or sediment ridges, which can produce sequences of detections that satisfy the consecutive observation criteria, and would have to be addressed through improved classifier performance~\cite{cailei_joe2025}. 

\subsection{Catenary-bounded Search Area}

The maximum displacement of a subsea cable at any given distance from an observed location can be approximated using catenary calculations~\cite{Faltinsen1990}. This provides an upper bound to limit the area that an AUV needs to search if it loses track of the cable during a survey without over-constraining the search and risking leaving potential cable locations unexplored. In these situations, the AUV advances along the estimated cable route while performing a laterally oscillating (zigzag) search pattern constrained within the search envelope and centred on the estimated cable route. Since the magnitude and direction of the external loads that cause cables to displace are unknown, we consider the limiting of the cable minimum breaking load $T_{MBL}$ acting normal to the cable. Beyond this, the cable would fail and be detectable through established cable interrogation techniques.

\begin{figure}[!tb]
    \centering
    \includegraphics[width=1\linewidth]{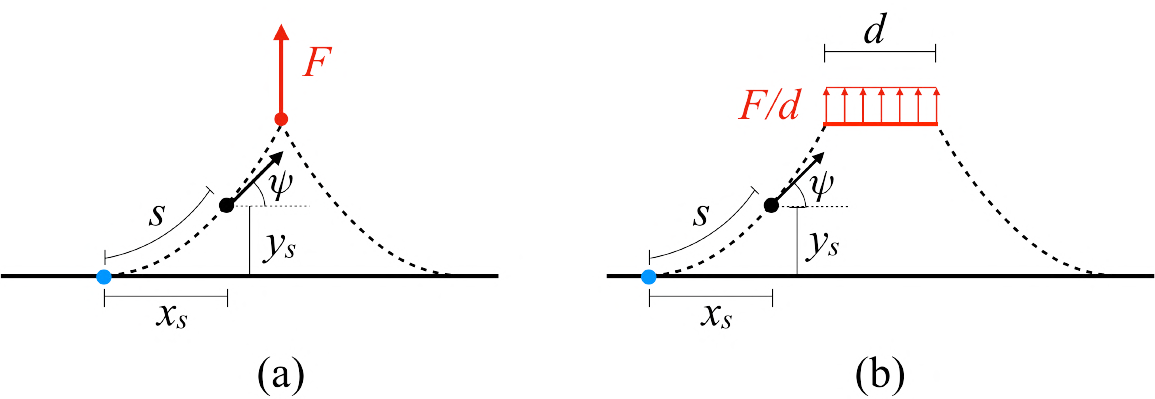}
    \caption{Top-down view of the cable lateral catenary (dotted line) resulting from (a) a point load, and (b) a uniformly distributed load, where both act normal to the as-laid cable (solid line), which achieves the largest lateral displacement.}
    \label{fig:cable_catenary}
\end{figure}

Fig. \ref{fig:cable_catenary} illustrates the effect of a point load $F$ acting normal to the cable (e.g. corresponding to the cable being trawled or dragged by an anchor), or distributed over some distance d (e.g., corresponding to a landslide). Since cables transfer tension along their arc, disturbed sections follow a catenary profile defined by the magnitude of the external load, the cable lay-tension and reactive forces (e.g., seabed friction). The distance along the catenary $s$, is measured from the point where the cable first deviates its position due to the external force (i.e., from the blue point in the figure). Distances $x_{s}$ and $y_{s}$ are the projections of the catenary along and normal to the undisturbed configuration with $\psi$ defined relative to it.

\begin{figure}[!tb]
    \centering
    \includegraphics[width=0.8\linewidth]{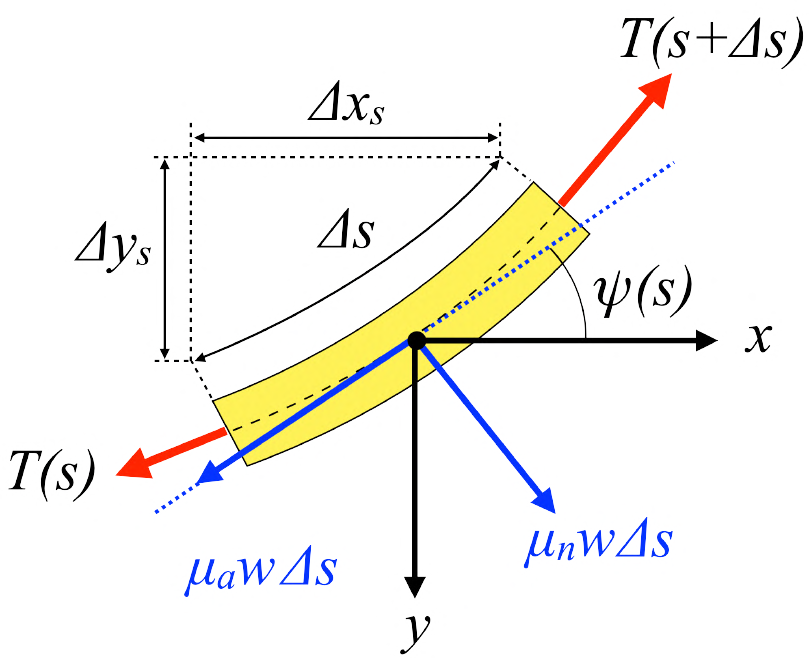}
    \caption{Plan view free-body diagram of a cable element. Friction between the cable and the seafloor generates reaction forces that cause the cable tension to vary along the arc length. These reaction forces depend on the immersed weight of the cable segment, $w\Delta s$, where $w$ is the immersed weight per unit length, and on the axial and normal friction coefficients, $\mu_a$ and $\mu_n$.}
    \label{fig:cable_catenary_forces}
\end{figure}

Friction between the cable and the seafloor opposes the external load, eventually balancing them at which point the cable remains undisturbed. Fig. \ref{fig:cable_catenary_forces} shows the free-body force diagram for a short catenary section of length $\Delta s$. The cable is assumed to be inextensible, with a minimum bend radius that is negligible compared to the load induced curvature; both of which are reasonable given that unarmoured communication cables have a minimum bend radius 1\,m and sufficient axial stiffness to protect fibres from tension during cable laying. We assume that the hydrodynamic drag is negligible due to the small cable diameters ($<3\,cm$) and currents in deep water (typically $<0.2\,m/s$) \cite{MCCAVE20171}.  Since the reactive forces (friction) are non-restoring, the cable remains displaced after the load is removed.

The catenary profile is determined by the equilibrium of forces acting on each cable element, where tension in the $x$ and $y$ directions are:

\begin{align}
\Delta T_{x}(\psi) &= w\Delta s(\mu_{n}\sin\psi - \mu_{a}\cos\psi) \label{eq:3_31} \\
\Delta T_{y}(\psi) &= w\Delta s(\mu_{n}\cos\psi + \mu_{a}\sin\psi).\label{eq:3_32}
\end{align}

\noindent Here $w$ is the unit length immersed cable weight and $\mu_{a}$, $\mu_{n}$ are the axial and normal seabed friction coefficients. Although these parameters can be modelled as variables of the system based on available geotechnical data (potentially interpreted from images), for most cables the effect of sediment properties on the friction coefficients is minor. This is because a low sediment strength (or friction) leads to higher embedment, which compensates by adding passive resistance laterally and a wedging effect axially~\cite{davidwhite2010}. Conservative search envelopes (i.e., larger) correspond to large friction coefficients, which results in the tightest catenary. At $s=0$, the cable is undisturbed with $\psi=0$, where the $T_{x}=T_{0}$ and $T_{y}=0$, respectively. The initial cable lay tension ($T_{0}$) is typically known determined by the immersed weight of the suspended cable during lay operations, which depends on $w$ and seafloor depth. The catenary profile is determined by numerically integrating the tension components $\Delta T_x$ and $\Delta T_y$ along the arc length $s$ from the undisturbed point. Cables with the same immersed weight, friction coefficients and lay tensions displace along identical profiles up to the point where the reaction force cancels the external load, meaning that the rate of growth of the search envelope is identical regardless of the magnitude of the external disturbance up to the limit:

\begin{equation}
T_{i} = \sqrt{T_{xi}^{2} + T_{yi}^{2}} < T_{MBL}. \label{eq:3_42}
\end{equation}

\begin{figure}[!tb]
    \centering
    \includegraphics[width=1\linewidth]{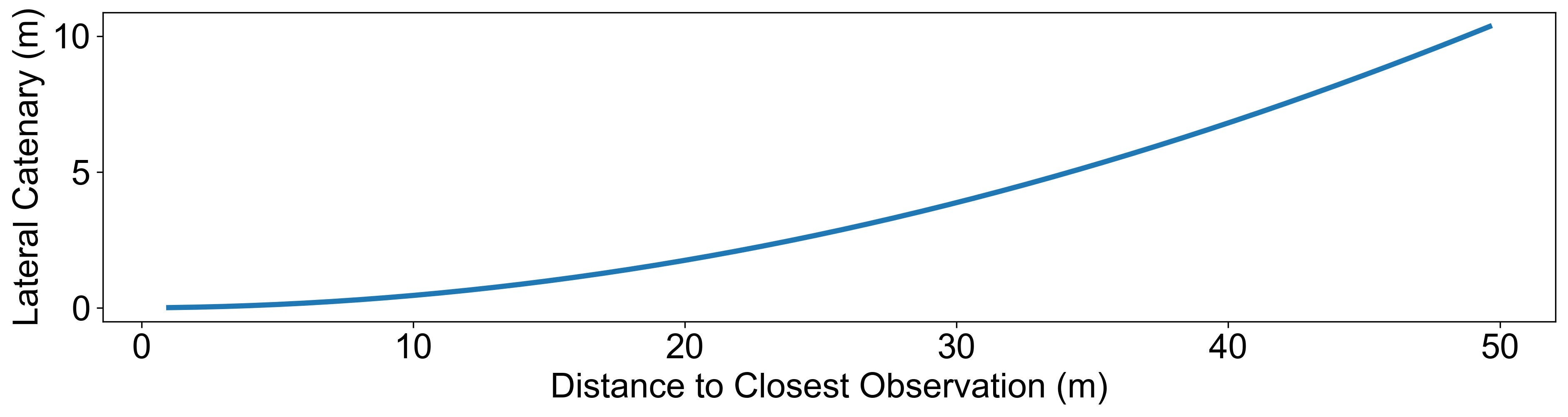}
    \caption{Search boundary curve for a subsea cable tracking mission with limits $\sigma_{\min}=0$\,m and $\sigma_{\max}=10$\,m. The curve is defined by the catenary projections $x_{si}$ and $y_{si}$, precomputed from the cable’s physical parameters.}
    \label{fig:catenary_curve}
\end{figure}

\noindent The cable's catenary can be calculated based on modelled physical parameters by assuming the search envelope starts at the nearest observed cable $\mathbf{m}_{j}$. Fig. \ref{fig:catenary_curve} illustrates the catenary limited search where $y_{si} = \sigma_i$ varies with distance $\mathbf{x}_{i}$ from $\mathbf{m}_{j}$. In practice, it can be useful to bound the maximum deviation from the cable route to avoid the robot navigating too far from the operating region in cases where it is unable to detect the cable for an extended period. This can be implemented by setting an upper limit on the search space $\sigma$ to the initial route uncertainty $\sigma_{max}$.

\subsection{Cable Search}

\begin{figure}[!tb]
    \centering
    \includegraphics[width=0.9\linewidth]{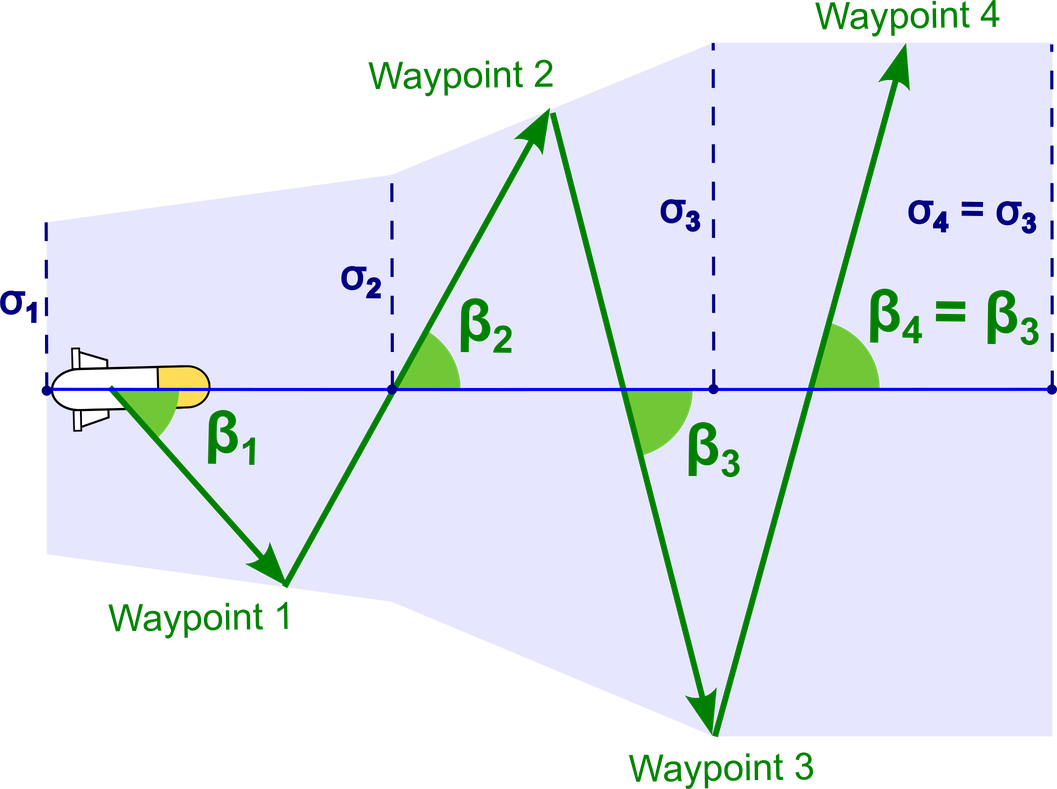}
    \caption{Illustration of the zig-zag search pattern for locating a subsea cable within a bounded search area. The search boundaries expand from the last observed cable location, starting at $\sigma_{\min}=\sigma_1$ and increasing to a maximum of $\sigma_{\max}=\sigma_3=\sigma_4$. Search waypoints are generated at angles $\beta_i$ relative to the cable coordinate frame, beginning at the AUV’s current position and extending to the boundary edge. The angles $\beta_1,\beta_2,\beta_3,\beta_4$ scale linearly with the boundary values $\sigma_1,\sigma_2,\sigma_3,\sigma_4$. Since $\sigma_1 < \sigma_2 < \sigma_3 = \sigma_4$, the corresponding angles satisfy $\beta_1 < \beta_2 < \beta_3 = \beta_4$.}
    \label{fig:cable_search_journal}
\end{figure}
Once the cable search boundary is defined, the AUV performs a zig-zag search pattern to locate the cable within the catenary-bounded search area~\cite{ut2025}. Initially, every cable node is assigned a search boundary of $\sigma_{max}$, which is typically 10\% of the seafloor depth. Once a section of cable is observed, the search boundary is updated using a catenary curve based on its distance to the last observation. Observed cables are assigned a minimum search boundary of $\sigma_{min}$, which is based on the AUV's navigational uncertainty. 

Fig. \ref{fig:cable_search_journal} illustrates the growth of the search boundary from $\sigma_{min}$ up to $\sigma_{max}$. The zig-zag search angle linearly scales with the search boundary range from $\beta_{min}$ to $\beta_{max}$. If the AUV exceeds $\sigma_{max}$ by more than a threshold distance (e.g., due to an incorrect cable tracking vector), it returns back to the search area travelling normal to the estimated cable route map.

After reaching the end of the planned cable survey, the updated cable route map and associated search boundaries can be used to initialise subsequent follow-up surveys along the same cable route. If observations were made, a follow-up survey begins with a more accurate cable map with a more tightly constrained search area, which increases the effectiveness of subsequent surveys.

\section{Experiment Setup}
\subsection{Field Trials and Parameter Selection}
\begin{figure}[!tb]
    \centering
    
    \begin{minipage}[b]{\linewidth}
        \centering
        \includegraphics[width=\linewidth]{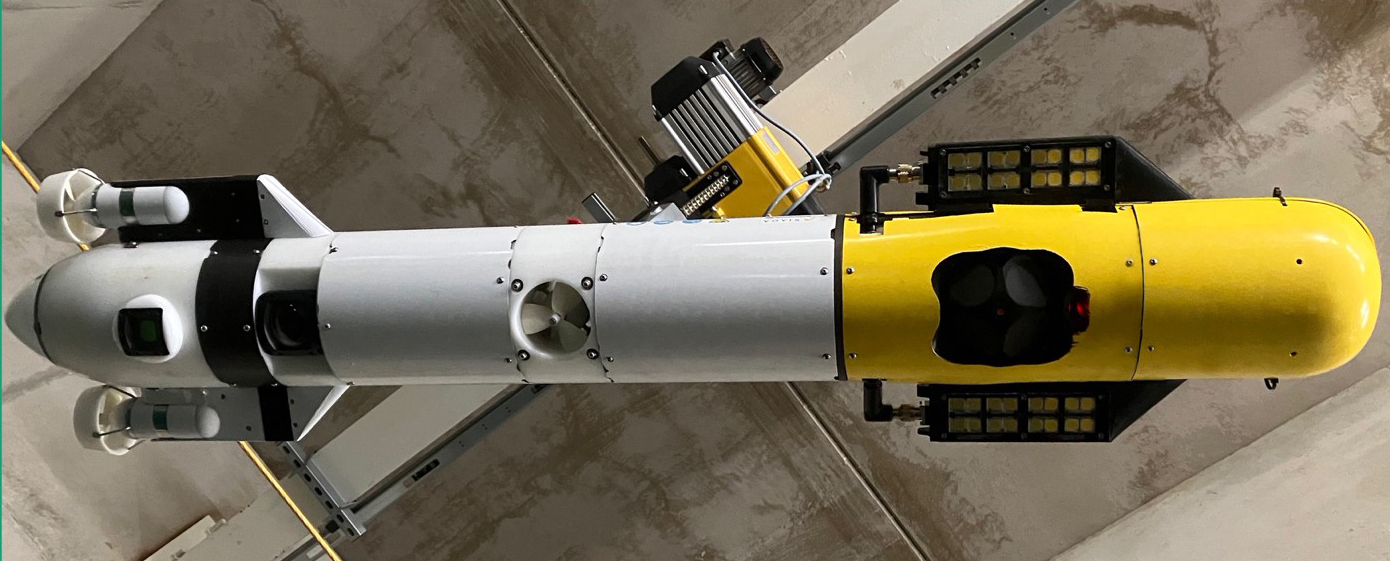}        
    \end{minipage}
    
    \vspace{0.5em} 
    
    \begin{minipage}[b]{\linewidth}
        \centering
        \includegraphics[width=\linewidth]{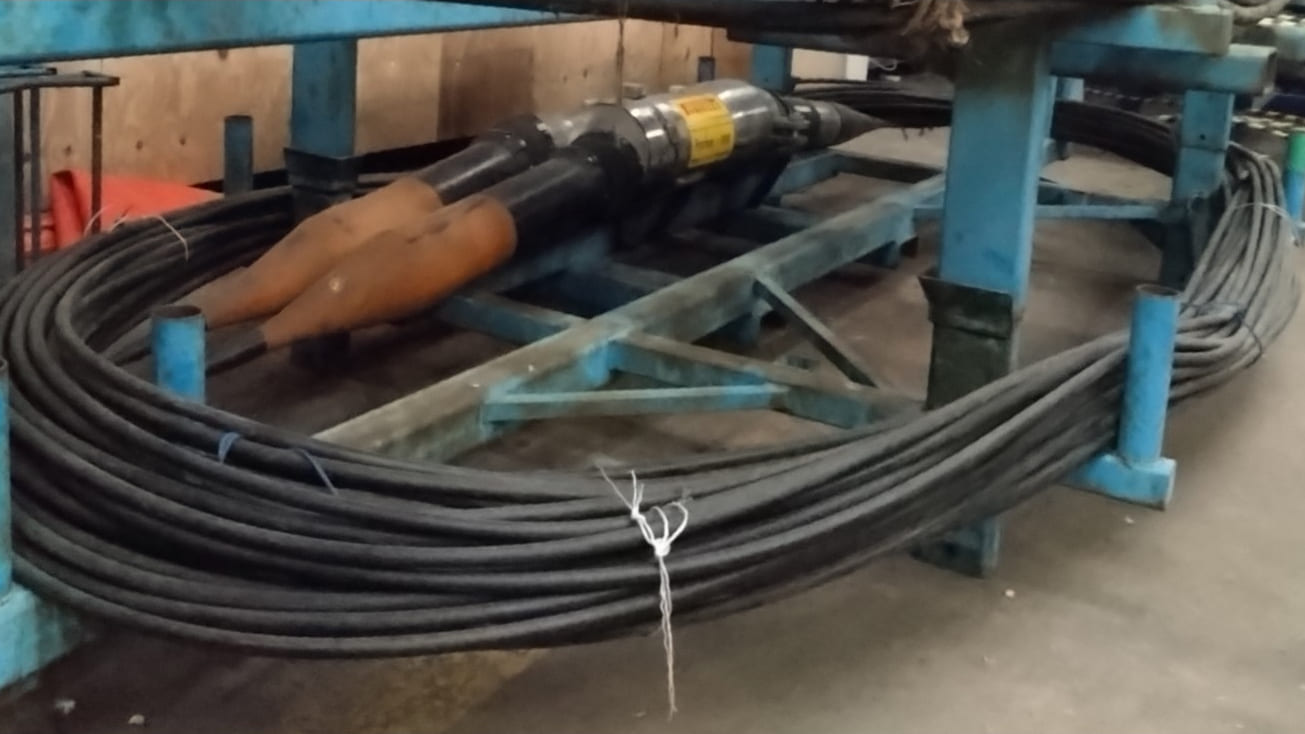}        
    \end{minipage}
    
    \caption{Top: Smarty200 is a 2\,m, 70\,kg AUV rated to 200\,m depth. It is equipped with a downward-looking camera with strobe illumination and a terrain-profiling laser sheet, a DVL-aided inertial navigation system, an acoustic transponder for positioning, and a GPS antenna for surface localisation. Bottom: A section of decommissioned subsea communication cable of diameter of 27\,mm that was used during the trials.}
    \label{fig:field_trial_setup_1}
\end{figure}

\begin{figure}[!tb]
    \centering
    \includegraphics[width=1\linewidth]{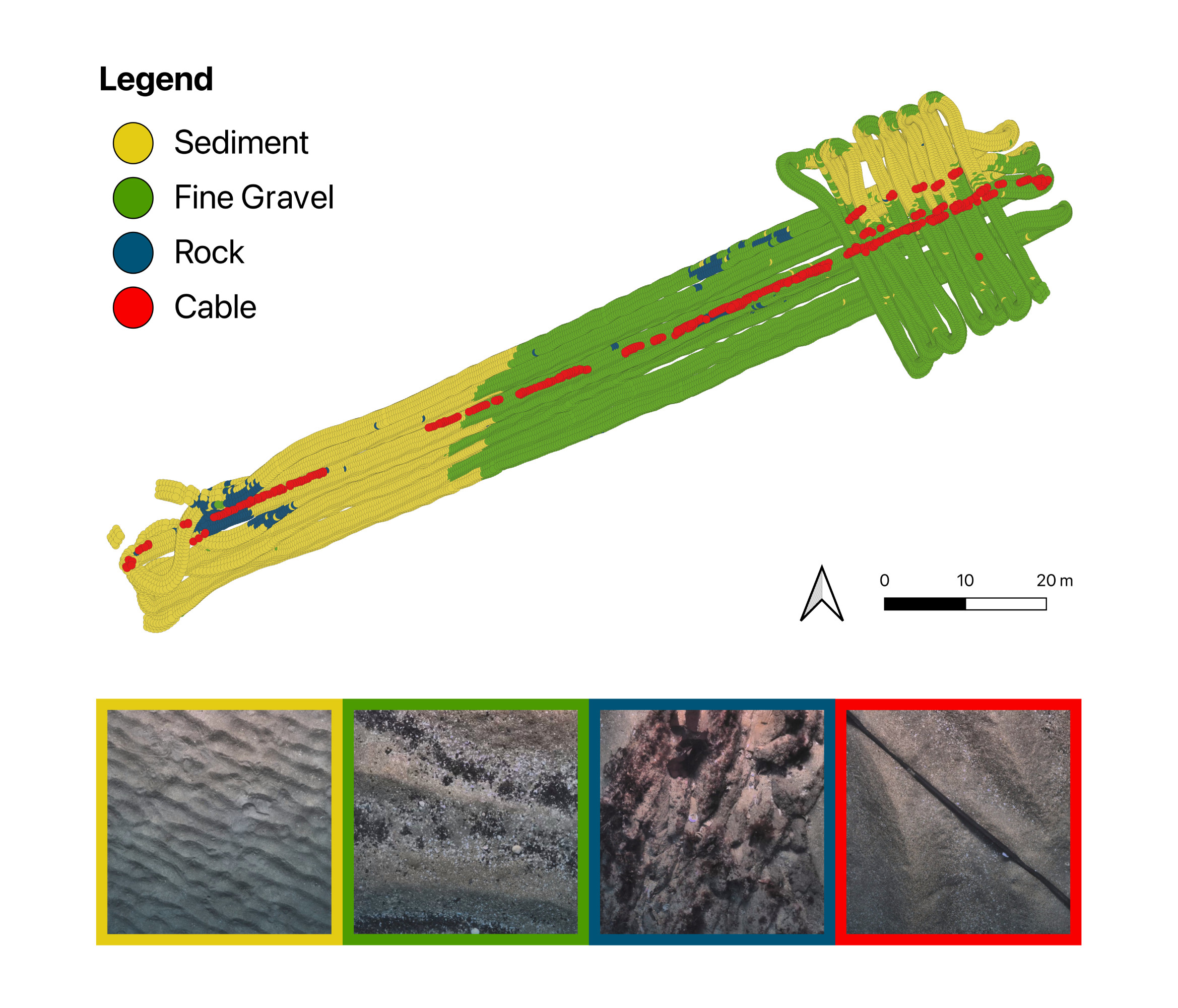}
    \caption{Seafloor substrate distribution at the test site. A dense lawnmower survey was conducted to gather images of the site once the cable was laid. Images were classified into four distinct categories offline following the method described in~\cite{geoclr_2022}. The panels in the figure show representative images of each class: Sediment, Fine Gravel, Rock and Cable.}
    \label{fig:lawnmower_labels}
\end{figure}

We demonstrated our method using the University of Southampton's Smarty200 AUV, shown in Fig.\,\ref{fig:field_trial_setup_1}(a), which was deployed off Cawsand Bay, UK, to survey a 27\,mm diameter decommissioned deep-sea communication cable cut from the reel in Fig.\,\ref{fig:field_trial_setup_1}(b). The cable's parameters are given in Table \ref{tab:cable_parameters}. 
The cable was laid in two 10\,m and 100\,m long sections that were laid in-line with a 10\,m gap to simulate partial burial. For practical reasons, the sections were connected using a white nylon rope. The seafloor was flat with a depth of $12\,m$ and substrates varying between sediment, rock and gravel. The site was mapped using the AUV after the cable was laid and is illustrated in Fig.~\ref{fig:lawnmower_labels}. The seafloor classes in the figure were determined offline using semi-supervised learning~\cite{geoclr_2022}.

\begin{table}[!b]
    \centering
    \caption{Physical parameters of the subsea cable}
    \begin{adjustbox}{max width=\textwidth}
    \begin{tabular}{|c|c|}
        \hline
        Parameter & Value \\
        \hline
        $T_{MBL}$ (kN) & 200 \\
        Lay tension (kN) & 1.54 \\
        Diameter (mm) & 27 \\
        Wet weight (N/m) & 12.7 \\
        Seabed friction [$u_{n}, u_{a}$] & [1.0, 0.1] \\
        \hline
    \end{tabular}
    \end{adjustbox}
    \label{tab:cable_parameters}
\end{table}

Smarty200 is a 200\,m depth rated hover-capable AUV equipped with a 12\,M pixel downward-facing camera with horizontal and vertical fields-of-view of $60^\circ$ and $46^\circ$, respectively. The AUV was programmed to perform cable inspection at a surge velocity of $0.2\,m/s$ while maintaining a constant altitude of $2$\,m, which translates to an image footprint of $2.3 \times 1.7$\,m at approximately 0.6\,mm resolution. The AUV estimates its pose using an Extended Kalman Filter (EKF) that fuses a DVL-aided inertial navigation system (Sonardyne SprintNav Mini) with USBL position updates from a surface vessel. The EKF position uncertainty sets a lower bound on the cable observation covariance $R_{jk}$. Although the framework can also directly use real-time EKF uncertainty estimates, during trials this was set to a fixed conservative value corresponding to $3$\,m position uncertainty, which is reasonable given the shallow depth of operations and short duration of each experiment. 

Preliminary trials showed that the cable detection model can process, on average, one image every $4.5$ seconds using Smarty200's on-board CPU. Although significantly faster rates can be achieved using the AUV's onboard Jetson Xavier GPU (0.3\,s), this was not implemented during the trials. At $0.2\,m/s$ forward speed, the AUV travels up to $0.9\,m$ between image frames. The tracking look-ahead distance $L_{wp}$ was set between $6\,m$ and $12\,m$ during our trials, allowing the AUV to follow tracking waypoints for $30-60$ seconds and process $6-12$ images before starting a search pattern if the cable is not detected during this time. The consecutive observation criteria was set at $d_{obs}=2\,m$ and $N_{obs}=3$ to allow at least two image frames to be processed so that the cable detection model could process at least two images before modifying the graph.

The prior cable route map was provided with a $7.5$\,m lateral offset from the actual cable route, which was determined using four acoustic transponders that were fitted along the cable length. The prior and actual cable map are shown in the first panel of Fig.~\ref{fig:follow_up_survey}. The initial search boundary is set at $\sigma_{max}=14.14$\,m, which is approximately $6$ times the camera swath, which can shrink down to a minimum of $\sigma_{min}=3$\,m ($1.3$ times the camera swath). While both the offset and uncertainties used are small, they are reasonable given the purpose of testing the approach over the short distance of cable available for the trials. This initial offset simulates the shifts in the subsea cable map position. The initial search boundary simulates the camera's limited field of view, which forces the AUV to search for the cable within the bounded area. The minimum search boundary ensures that the AUV cannot see the entire search area within its field of view, leaving room for the AUV to zigzag inside the search area. The cable search angle range was set between $\beta_{min}=30^\circ$ to $\beta_{max}=50^\circ$ based on simulations described in our previous work~\cite{udt2024} .


\subsection{Survey Efficiency}

\begin{figure}[!tb]
    \centering
    \includegraphics[width=\linewidth]{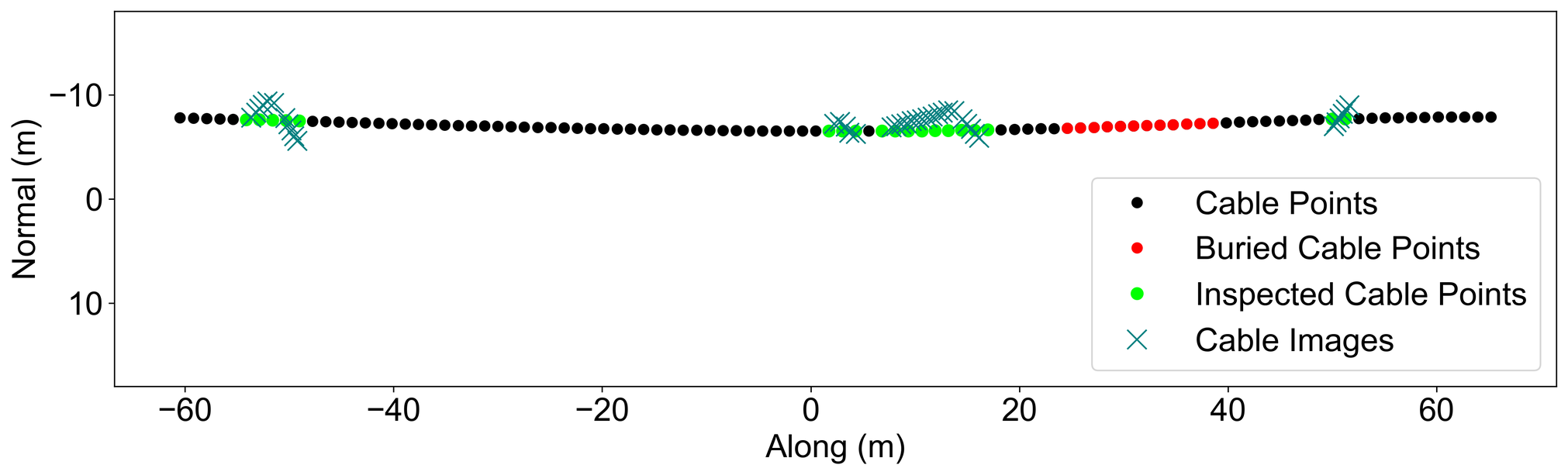}
    \caption{Inspection efficiency is evaluated by sampling the true cable route into $N=100$ evenly spaced points. Image locations containing visible cable ($\times$) are associated with the nearest cable node (black circles) along the $x_c$ axis. These nodes are marked as inspected (green circles) for efficiency calculation. Buried cable nodes (red circles) are not observable and are excluded from the total count of visible cable points.}
    \label{fig:inspection_efficiency}
\end{figure}

The survey efficiency is defined as the product of inspection efficiency ($\eta_{inspection}$) and distance efficiency ($\eta_{distance}$):

\begin{equation}
    \eta_{survey}=\eta_{inspection} \cdot \eta_{distance}
\end{equation}

\noindent where the inspection efficiency evaluates the proportion of the visible cable successfully captured by the AUV's camera, illustrated in Fig. \ref{fig:inspection_efficiency} where the actual cable route is discretized into $N$ points that are flagged as inspected when they fall into an image footprint with a visible section of cable. 
Buried sections are excluded from the total number of visible points.

\begin{equation}
    \eta_{inspection}=\frac{\rm Total\ cable\ points\ inspected}{\rm Total\ visible\ cable\ points}.
\end{equation}

\noindent The distance efficiency compares the true length of the cable against the total distance travelled by the AUV

\begin{equation}
    \eta_{distance}=\frac{\rm Total\ cable\ length}{\rm Total\ distance\ travelled}.
\end{equation}

\section{Results and Discussion}

\subsection{Dive Analysis}


We compare the results from four dives (Table \ref{tab:dive_specifications}), three of which used the proposed real-time cable tracking method and the other a dense lawnmower survey pattern. The cable route was sampled to $N=100$ evenly-spaced nodes, resulting in a node spacing of $1.2\,m$. This is comparable to the AUV's image footprint in the forward direction, which provides a practical upper bound on useful cable route resolution. A practical upper bound on cable route node intervals would match the initial route uncertainty, as beyond this the catenary profile would be undersampled. A node spacing of $10\,m$, would result in a graph consisting of $10,000$ cable nodes ($20,000$ variables) for a long-range AUV imaging missions (on the order of $100\,km$~\cite{adrian_tfr2025}). This remains computationally manageable using efficient graph solvers such as iSAM2~\cite{kaess2012isam2}. If higher route resolutions are needed, the cable route map can be split into shorter overlapping segments without major performance loss.

\begin{figure}[!htb]
    \centering
    \includegraphics[width=\linewidth]{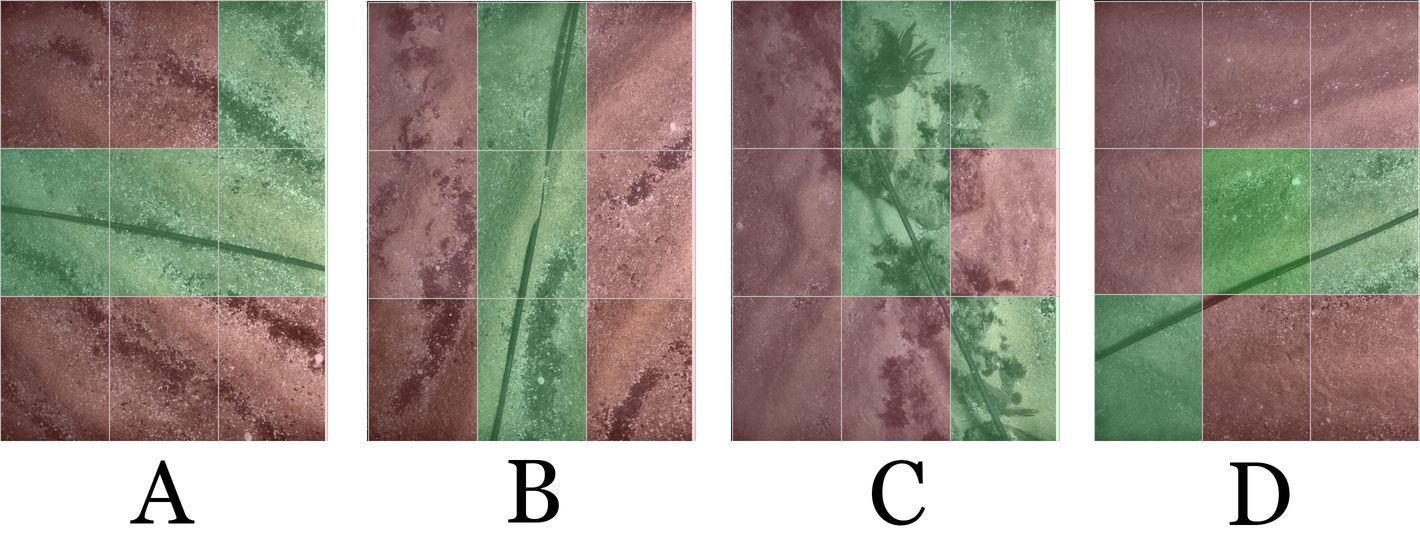}
    \caption{Results of real-time cable detection in a $3\times3$ grid, where green tiles indicate cable detections and red tiles indicate no detection. Image A shows correct detection of the cable, with an additional false positive caused by gravel within sand ripples. Image B shows accurate detection despite gravel and partial burial. Image C exhibits both false positive and false negative detections due to complex rock and algae. Image D shows a missed detection in the centre tile of the bottom row, where the cable occupies only a small portion of the tile.}
    \label{fig:cable_detections}
\end{figure}
\begin{figure}[!tb]
    \centering
    \begin{minipage}[b]{\linewidth}
        \includegraphics[width=1\linewidth]{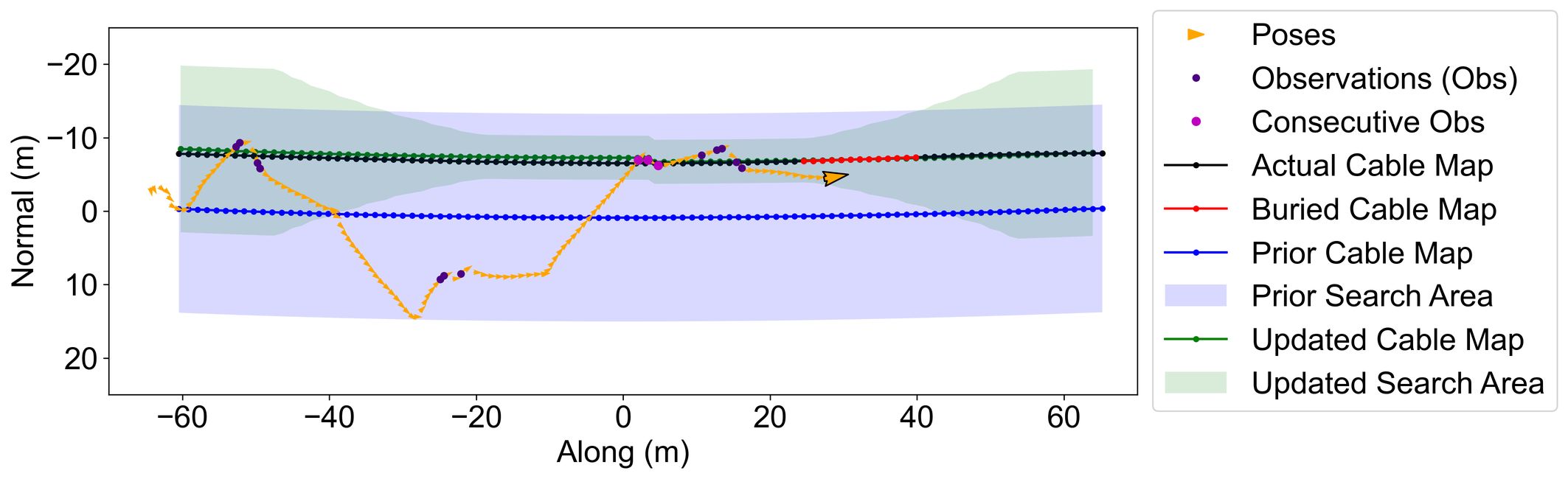}
	\end{minipage}
    \begin{minipage}[b]{\linewidth}
		\centering
        \includegraphics[width=1\linewidth]{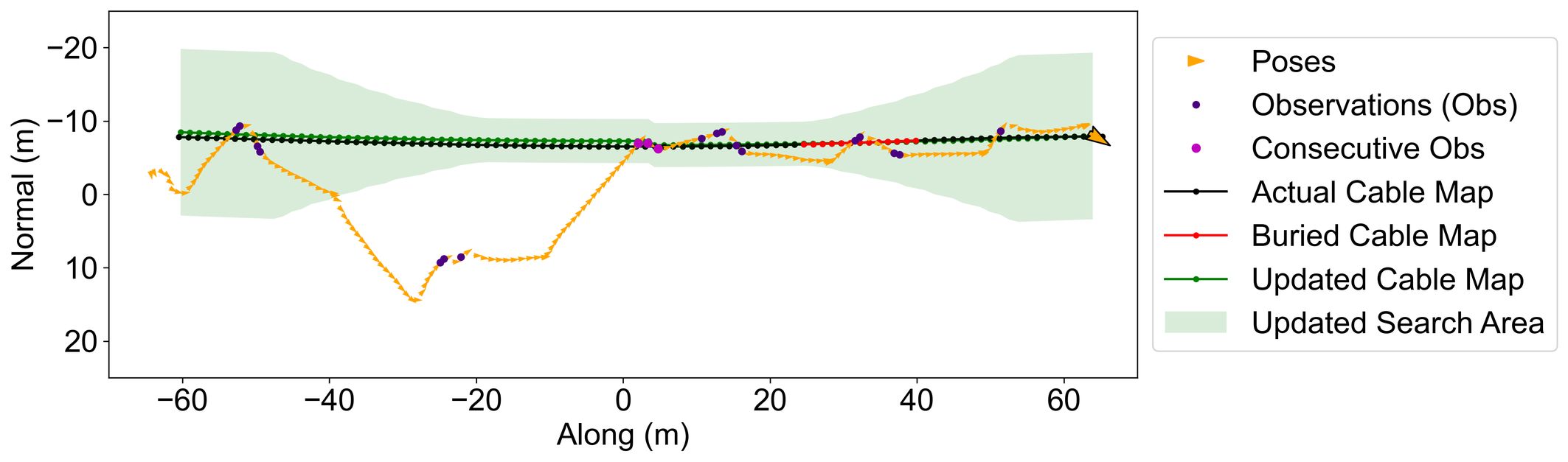}
	\end{minipage}
    \begin{minipage}[b]{\linewidth}
		\centering
        \includegraphics[width=1\linewidth]{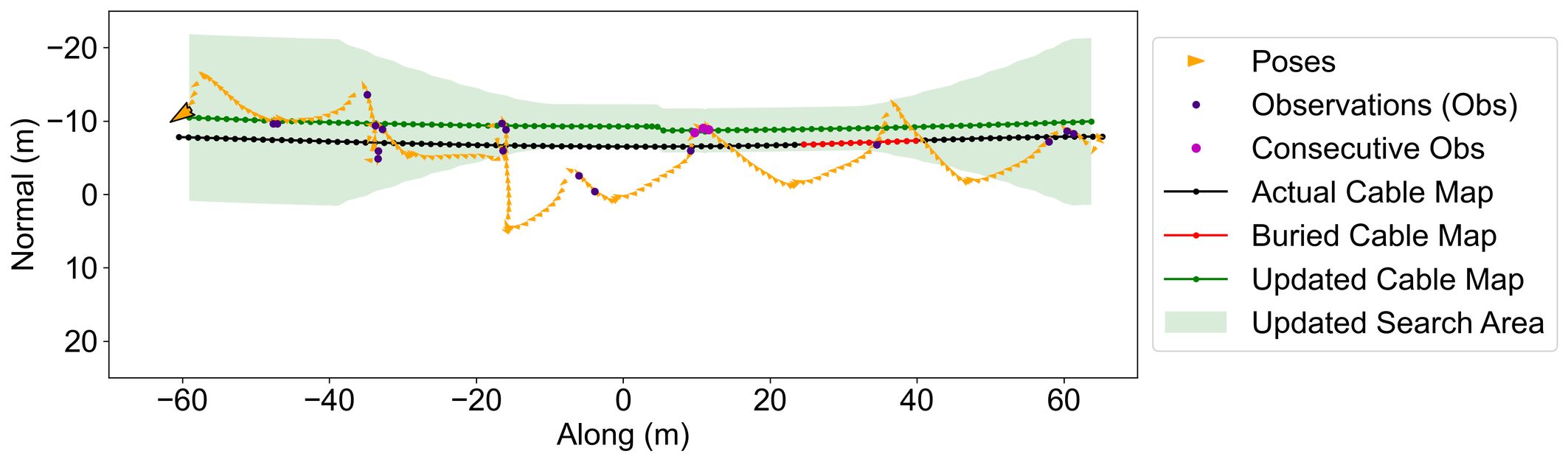}
	\end{minipage}
    \caption{Dive 1 stages. The top plot shows the 
    cable map update moved the entire cable map (green dotted lines) from the initial cable map (blue dotted lines) offset $7.5$\,m from the true cable (black dotted lines) to the top of the three consecutive observations (violet dots) at $x_c=0, y_c=-7.5$\,m. The search area was also updated from an initial boundary of $\sigma_{max}$ (blue area) to a catenary curve (green area) growing in both the positive and negative directions $along$ axis. The middle plot shows the updated cable map from the initial survey, used as the starting map for the follow-up survey, with all poses and observations reset and the AUV travelling in the reverse direction. The bottom plot shows the end of the follow-up survey, where the cable map is further shifted along the negative normal axis to align with new observations.}
    \label{fig:follow_up_survey}
\end{figure}

During cable tracking, the AUV starts searching for the cable within the initial search area (see Fig.~\ref{fig:follow_up_survey}). Images are taken at 1\,Hz with cable detection running on the latest image, where observations are determined based on the AUV pose when the corresponding image was taken. Examples of cable detection over various substrates are shown in  Fig. \ref{fig:cable_detections}.

\begin{table}
    \centering
    \caption{Specification of the three complete cable tracking dives during the field trials}
    \begin{tblr}{
      width = \linewidth,
      colspec = {Q[77]Q[270]Q[270]Q[192]},
      cells = {c},
      hlines,
      vlines,
    }
    Dive & Date \& Time & Initial Direction & $L_{wp}$  (m)\\
    1 & 20250326\_151609 & East to West &  12\\
    2 & 20250326\_164956 & West to East &  12\\
    3 & 20250327\_095310 & West to East &  6
    \end{tblr}
    \label{tab:dive_specifications}
\end{table}


The top figure of 
Fig.~\ref{fig:follow_up_survey}
illustrates how the cable map position and search boundary are updated after consecutive observations satisfied the cable optimisation conditions. The updated search boundary expands from the last consecutive observation positions following the cable's catenary curve, growing from $\sigma_{min}$ up to $\sigma_{max}$. Out of all the cable observations in the figure, only the consecutive observations near $x_c=0, y_c=-7.5$\,m satisfied conditions for cable route optimisation. The other observations were used to track the cable, but did not update the cable route map as they were not consecutive. While several of these non-consecutive observations were on top of the actual cable, the group of three observations near $x_c=-25\, y_c=10$\,m were false positives. Although these triggered erroneous cable tracking up to the look ahead distance, the cable route map was not updated and the AUV was able to resume its search to relocate the cable. 



After reaching the end of the cable map, a map update is performed to finalise the cable map positions and search boundaries. During the field trials, each successful mission was continued with a follow-up survey using the updated cable map from the initial survey. The follow-up survey was performed immediately after the initial survey in the opposite direction. Fig.~\ref{fig:follow_up_survey} shows the cable map after a graph optimisation, the transition to the follow-up survey, and the end of the follow-up survey. On the return leg, the robot detects a cable tracking vector that sends it away from the actual cable at $x_c=-8, y_c=-2$\,m, which sends the robot away from the cable. However, at $x_c=-18, y_c=5$\,m, the robot exceeds the safety threshold (set at 10\,m) beyond the route uncertainty beyond which the AUV returns orthogonally to the cable route to recover cable tracking.

The inspection efficiencies of the three completed dives are shown in Table \ref{tab:inspection_efficiency}, where the combined inspection efficiency captures results after both the initial and follow-up surveys. 
Dive 3 started with the lowest inspection efficiency in its initial survey, but achieved the highest efficiency in its follow-up survey. This is partly due to the use of a shorter waypoint look-ahead distance $L_{wp}=6\,m$, which reduces the distance travelled before starting to search after loss of tracking.


\begin{table}[!tb]
    \centering
    \caption{Inspection efficiency of the initial, follow-up, and their combined results across three cable tracking dives and a lawnmower survey. For the cable tracking dives, distance and survey efficiencies are computed using the combined surveys.}
    \begin{tblr}{
      width = \linewidth,
      colspec = {
        c
        X[1.12,c,m]
        X[1.5,c,m]
        c
        X[2.23,c,m]
        X[2.23,c,m]
      },
      cells = {c,m},
      cell{1}{1} = {r=2}{},
      cell{1}{2} = {c=3}{},
      cell{1}{5} = {r=2}{},
      cell{1}{6} = {r=2}{},
      hlines,
      vlines,
    }
    Dive & Inspection Efficiency (\%) &  &  & Distance Efficiency (\%) & Survey Efficiency (\%)\\
     & Initial & Follow-up & Combined &  & \\
    1 & 21.59 & 36.36 & 52.27 & 26.62 & 13.91\\
    2 & 29.55 & 46.59 & 57.95 & 21.18 & 12.27\\
    3 & 10.23 & 54.55 & 59.09 & 34.94 & 20.65\\
    Lawnmower & 100 & - & - & 8 & 8
    \end{tblr}
    \label{tab:inspection_efficiency}
\end{table}

The lawnmower survey in Fig.~\ref{fig:lawnmower_labels} achieves 100\% inspection efficiency, but the distance efficiency is low at 8\%  because the AUV needs to covers the entire prior cable search range. The cable tracking dives reach 52--59\% combined inspection at distance efficiencies of 21--35\%, between 2.6 and 4.4 times higher than the lawnmower survey. 
However, approximately $41\%$ of the cable remained uninspected even in the best dive (Dive 3, combined). The main cause for this is the large number of false positive cable detections that divert the AUV off-route. One of the reasons for this was the relatively slow image processing rate,
which prevented false detection from being rapidly corrected by subsequent true positives detections. 

\subsection{Robustness Analysis}
\label{sec:robustness}

\begin{figure}[!tb]
    \centering
    \includegraphics[width=\linewidth]{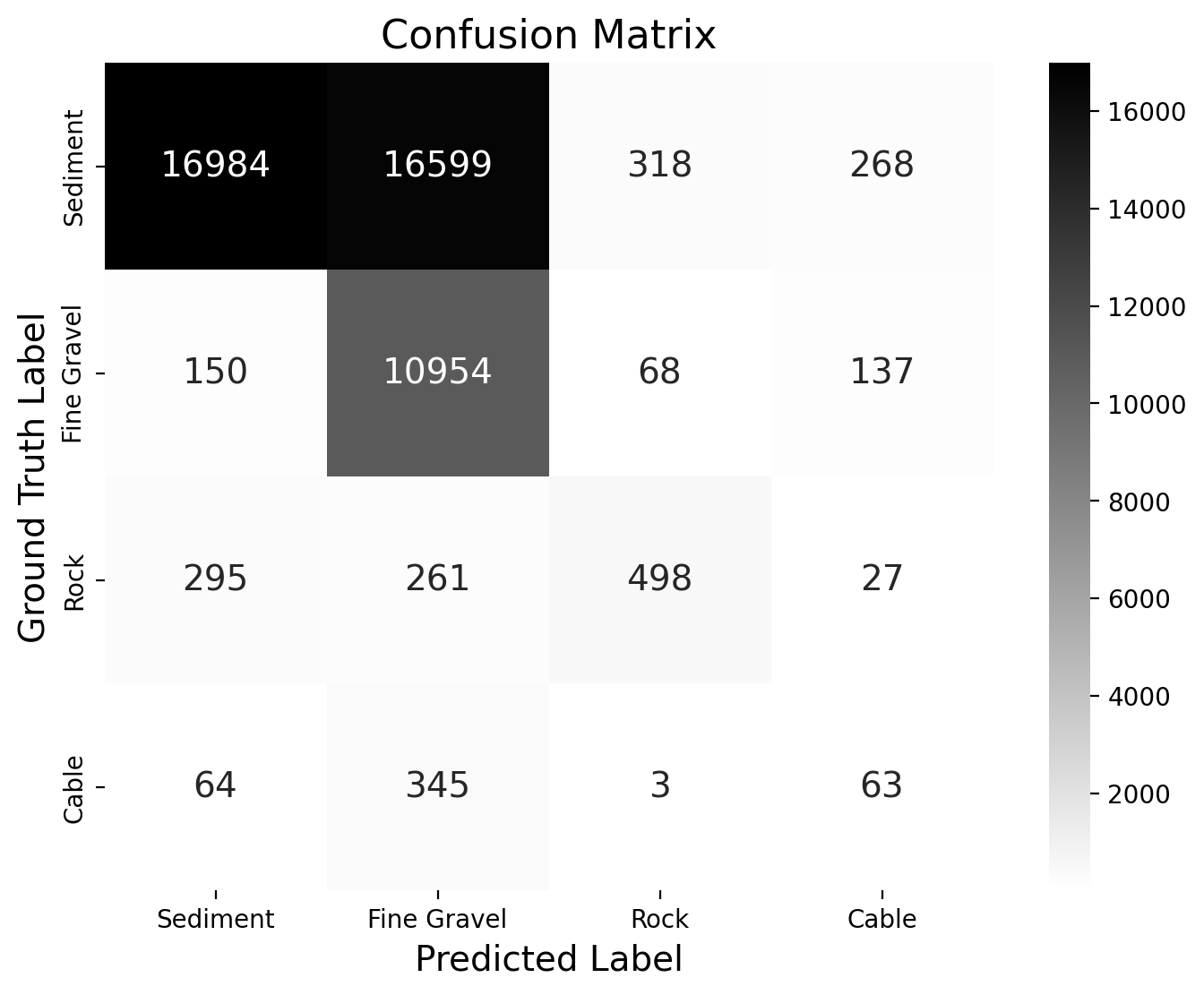}
    \caption{Confusion matrix for the GeoCLR classifier evaluated on the 47,034 images from the lawnmower survey (Fig.~\ref{fig:lawnmower_labels}). Rows show ground-truth labels while columns show predicted labels. The cable class has 475 ground truth samples out of 46,559 background images. The cable class has precision: 12.7\%, recall: 13.3\%, and F1-score: 13.0\%.}
    \label{fig:confusion_matrix}
\end{figure}

Fig.~\ref{fig:confusion_matrix} characterises the cable classifier's performance across the four substrate types at the trial site. For navigation, the most
consequential metric is the false positive (FP) rate where a non-cable image is classified as cable, generating spurious waypoint that the AUV follows~\cite{auv2024}. 

Table~\ref{tab:fp_by_substrate} gives the FP cable detection rates calculated over different substrates. Rock has the highest FP rate at 2.5\%, where angular textures can resemble the cable's cylindrical profile. Sediment is lowest at 0.78\%, where its flat featureless surface provides the greatest contrast with the cable. Of the 475 cable ground-truth images in Fig.~\ref{fig:confusion_matrix}, the false negative (FN) rate is high, with 72.6\% misclassified as Fine Gravel, where Fig.~\ref{fig:lawnmower_labels} and Fig. \ref{fig:cable_detections} show the dark linear patterns formed by gravel that can confuse the cable detection model. 13.5\% where classified as Sediment, with 0.6\% classified as Rock. These results are partly skewed by the unbalanced number of images for each substrate type and the relatively small number of cable images. However, as described in our previous work, the method is less sensitive to FN than FPs, \cite{auv2024}.


\begin{table}[!b]
    \centering
    \caption{Per-substrate cable detection false positive rate, from the lawnmower survey confusion matrix (Fig.~\ref{fig:confusion_matrix}).}
    \label{tab:fp_by_substrate}
    \begin{tblr}{
      width = \linewidth,
      colspec = {Q[200]Q[192]Q[319]Q[212]},
      column{even} = {c},
      column{3} = {c},
      hlines,
      vlines,
    }
    Substrate & GT Images & Classified as Cable & FP Rate (\%)\\
    Sediment & 34,169 & 268 & 0.78\\
    Fine Gravel & 11,309 & 137 & 1.21\\
    Rock & 1,081 & 27 & 2.50
    \end{tblr}
\end{table}

The consecutive observation filter blocks isolated FPs but not consecutive ones that can be caused by rock ridges and striations visible in Fig.~\ref{fig:lawnmower_labels}, which can produce sequential FPs that satisfy our filtering conditions. Prior simulation using GeoCLR~\cite{auv2024} at a 2\% false positive and 25\% false negative rate achieved $90.3 \pm 4.7\%$ cable coverage. Field results came in at $52-59\%$, lower than that simulation figure. The gap is caused by the mixed substrates present at the field and the slower CPU-based inference used during the trials. GPU-accelerated inference was feasible on the AUV's onboard computer but was not deployed due to instabilities in the implementation. A higher processing rate would shorten the inter-detection travel distance and allow faster course corrections, increasing the relative proportion of cable images on which the classifier is applied. Furthermore, recent developments of Visual  Transformer (ViT) based detectors outperform  Convolutional Neural Network (CNN) models such as those used in GeoCLR on comparable seafloor datasets~\cite{cailei_joe2025}. These improvements would reduce both the FP and FN rates, which is expected to improved the overall performance of the system across all substrate types.

\section{Conclusion}


This work demonstrates that combining real-time visual tracking with graph-based optimisation enables robust subsea cable search and tracking despite uncertainty in prior route maps. By constraining the search space using physics-based catenary models, the method maintains efficient recovery from tracking loss while ensuring consistency with observations. Field trials on a $120$\,m cable showed up to $59$\% inspection coverage and survey efficiency up to $2.6$ times higher than a lawnmower survey. These results highlight the effectiveness of integrating perception, optimisation, and physical constraints for autonomous subsea cable inspection.


\balance  

\bibliographystyle{ieeetr}
\bibliography{references}

\end{document}